\documentclass[conference]{IEEEtran}
\IEEEoverridecommandlockouts
\usepackage{cite}
\usepackage{amsmath,amssymb,amsfonts}
\usepackage{algorithmic}
\usepackage{graphicx}
\usepackage{textcomp}
\usepackage{xcolor}

\usepackage{subfigure}
\usepackage{multirow}
\usepackage{makecell}
\usepackage{latexsym}
\usepackage{amsthm}
\usepackage{booktabs}
\usepackage{enumitem}
\usepackage[ruled,vlined,linesnumbered]{algorithm2e}

\usepackage{stfloats}
\usepackage{float}

\usepackage{listings}
\usepackage{siunitx}
\usepackage{tabularx}
\usepackage[table]{xcolor}
\usepackage{threeparttable} 

\def\BibTeX{{\rm B\kern-.05em{\sc i\kern-.025em b}\kern-.08em
    T\kern-.1667em\lower.7ex\hbox{E}\kern-.125emX}}
\begin{document}

\title{
 Diffusion-based 4D Trajectory Prediction and Distributed Control for UAV Swarms
\\
}

\author{Tianshun Li$^{1}$, Hongliang Lu$^{2}$, Haoang Li$^{3}$ and Xinhu Zheng$^{3, \ast}$ 
\thanks{$^\ast$ Corresponding author.}
\thanks{$^{1}$ Tianshun Li is with The Hong Kong University of Science and Technology (Guangzhou), China
        {\tt\small tli449@connect.hkust-gz.edu.cn}}%
        \thanks{$^{2}$Hongliang Lu is with the Southern University of Science and Technology, China and MoSense Technologies, China
        {\tt\small honglianglu36@gmail.com}}
        \thanks{$^{3}$Haoang Li and Xinhu Zheng are with The Hong Kong University of Science and Technology (Guangzhou), China
        {\tt\small \{haoangli,xinhuzheng\}@hkust-gz.edu.cn}}
}

\maketitle

\begin{abstract}
Accurate 4D trajectory prediction and closed-loop tracking are essential for Unmanned Aerial Vehicle (UAV) swarms to achieve safe and efficient operations in complex low-altitude environments such as urban airspaces, industrial sites, and indoor facilities. However, this task remains challenging due to intrinsic nonlinearity of UAV swarm dynamics and strict real-time constraints of swarm formation control.
To address these challenges, we propose a unified framework that couples coarse-to-fine trajectory forecasting with uncertainty-aware Distributed Nonlinear Model Predictive Control (DNMPC). Our approach features two key innovations: 1) a dimension-decoupled trajectory prediction module that reduces computational complexity by forecasting axis-wise motion, and 2) a diffusion-based residual dynamics refinement module that captures temporally correlated dynamic uncertainties. These refined predictions are then integrated into a DNMPC loop to ensure formation stability.
We also introduce a synchronized multi-scenario 4D UAV swarm dataset spanning six representative airspace scenarios. The dataset contains over \textbf{7,900} frames of synchronized three-UAV trajectories with frame-level annotations of speed intention and target sector.
Extensive experiments demonstrate that our approach outperforms state-of-the-art baselines, reducing trajectory tracking error by up to \textbf{10-15\%} and achieving sub-\textbf{0.07\,m} average tracking error in complex urban and industrial environments, while maintaining real-time inference speeds of 34 FPS (sub-30 ms latency) suitable for agile flight.
\end{abstract}

\section{Introduction}

Birds exhibit remarkable swarm intelligence to maintain compact formation in highly uncertain aerial environments; however, endowing UAV swarms with simultaneous capabilities remains a fundamental challenge~\cite{AeroDuo}. 
In recent years, UAVs have demonstrated strong potential in applications such as low-altitude logistics, surveillance, inspection, and indoor operations. 
When deployed as swarms in 4D airspace, UAVs must operate under significant temporal uncertainty arising from dense interactions.
To achieve compact formation maintenance, each UAV is required to arrive at its destination at the expected time point. But accurate 4D trajectory prediction for multiple UAVs becomes markedly more difficult, as modeling their joint spatiotemporal evolution requires reasoning over strongly coupled dynamics; the temporal dimension further exacerbates the difficulty of 4D trajectory prediction. Second, increased interaction density amplifies mutual collision risks, demanding predictive models that are compatible with coordinated control and formation maintenance. Third, additional sources of uncertainty, such as communication delay, downwash, and aerodynamic disturbances, introduce stochasticity that is difficult to model and propagate over time.
Taken together, maintaining compact formation remains challenging for UAV swarms to operate safely and simultaneously in complex low-altitude environments.

\begin{figure}[t]
    \centering
    \includegraphics[width=1\linewidth]{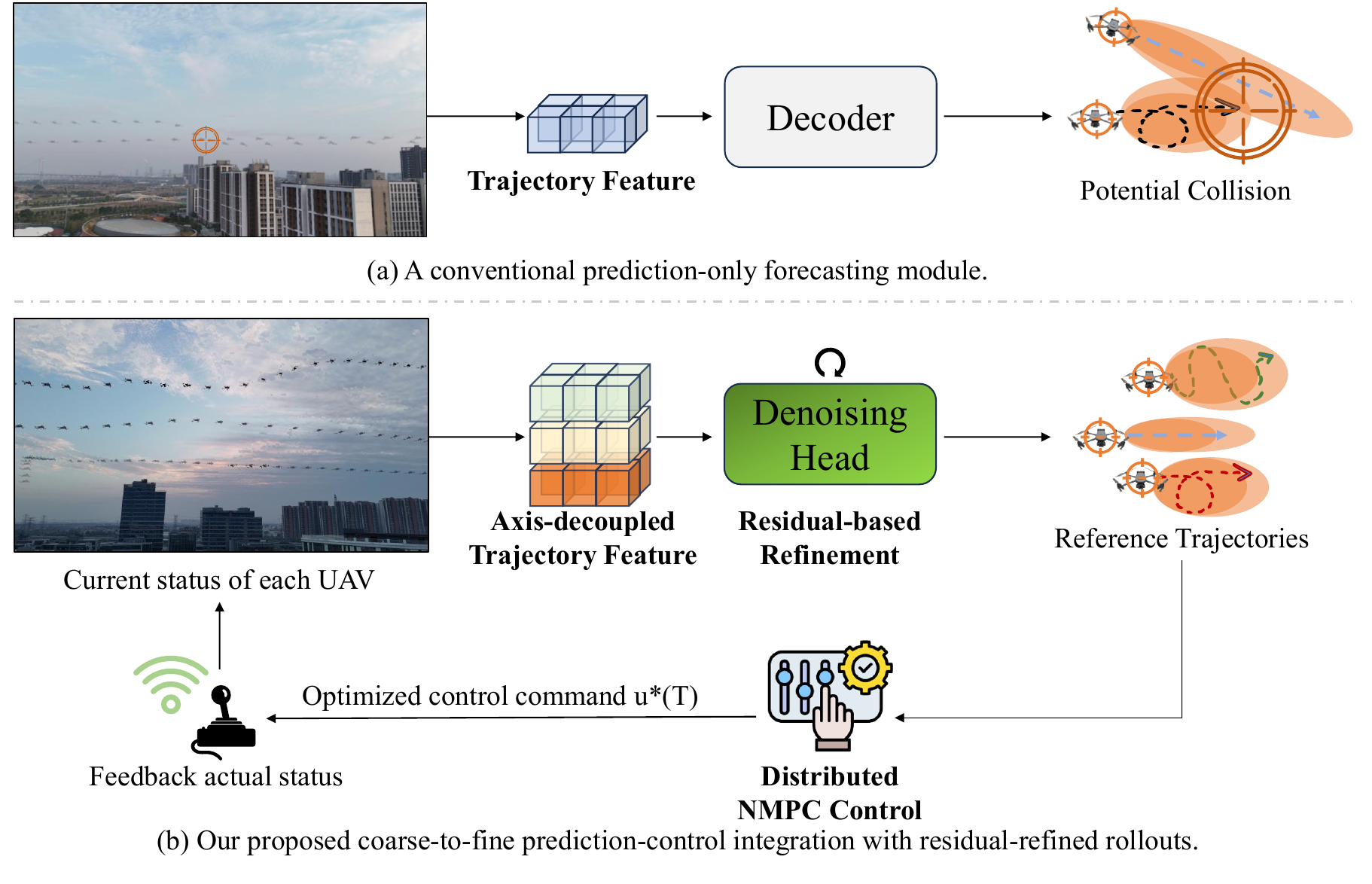}
    \caption{Conceptual comparison between a prediction-only forecasting module and prediction-control integration with residual-refined rollouts.
}
    \label{fig:figure1}
\end{figure}

Currently, most existing prediction approaches focus on 2D or 3D settings, either omitting control feedback or treating time as an auxiliary variable, as shown in Fig.~\ref{fig:figure1}(a). Such formulations are inherently limited in their ability to capture long-horizon temporal dependencies, and coordinated group behaviors that are fundamental to UAV swarms.
Moreover, although trajectory prediction for pedestrians and ground vehicles has been extensively studied, the advantages of these approaches do not readily translate into UAV swarms. These methods typically assume planar motion and weak inter-agent coupling, which are insufficient for aerial swarms, where formation maintenance, altitude variation, and aerodynamic coupling introduce additional complexity.

To address this problem, several model-based and data-driven approaches have been employed for UAV swarm 4D trajectory prediction; however, they still suffer from unstable performance~\cite{Yan2023CooperativeTA,pang2022adaptive}.
First, classical model-based methods rely heavily on faithful nominal dynamics and well-characterized uncertainty bounds, assumptions that are difficult to satisfy in real-world swarm flight due to nonlinear dynamics, aerodynamic disturbances, and communication delays.
Second, many learning-based trajectory prediction methods adopt deterministic formulations or simplified uncertainty structures, limiting their ability to represent temporally correlated disturbances such as airflow interaction and downwash effects.
Third, existing approaches often treat trajectory prediction and control as separate modules, making it difficult to propagate prediction uncertainty into downstream control decisions, especially under nonlinear dynamics, strong inter-agent coupling, and limited information exchange.
These limitations highlight the necessity of integrating uncertainty-aware 4D trajectory prediction with closed-loop control in a unified and scalable manner.

In this study, we propose a unified framework for 4D UAV swarm trajectory prediction and uncertainty-aware control, which jointly addresses spatiotemporal forecasting and closed-loop control under uncertainty. By explicitly coupling prediction and control, the framework enables reliable multi-UAV coordination under nonlinear dynamics, strong inter-agent interactions, and limited information exchange. As shown in Fig.~\ref{fig:figure1}(b), 
our framework decouples spatial axes for coarse prediction, refines temporally correlated residual dynamics via denoising, and ultimately embeds future-aware predictions into a DNMPC loop for closed-loop UAV swarm control.
The key contributions are summarized as follows:

\begin{itemize}
    \item Unified prediction-control framework: We propose a coarse-to-fine architecture that combines efficient axis-wise forecasting with diffusion-based residual modeling. This design effectively captures nonlinear swarm dynamics and temporally correlated disturbances without sacrificing real-time feasibility.
    \item Stability-Oriented DNMPC Integration: We theoretically analyze and empirically validate the integration of diffusion-based predictions into a DNMPC scheme. Compared with prediction-only forecasting modules that are typically evaluated without explicit control-oriented residual compensation, our framework refines coarse forecasts using diffusion-based residual dynamics and feeds the refined rollouts into a DNMPC loop.
    \item Dataset: We release a diverse multi-scenario dataset featuring synchronized 4D trajectories (3D position + time) with frame-level annotations of speed intention and target sectors, addressing the scarcity of realistic benchmarks for close-proximity swarm operations.
\end{itemize}

\begin{figure*}
    \centering
    \includegraphics[width=1\linewidth]{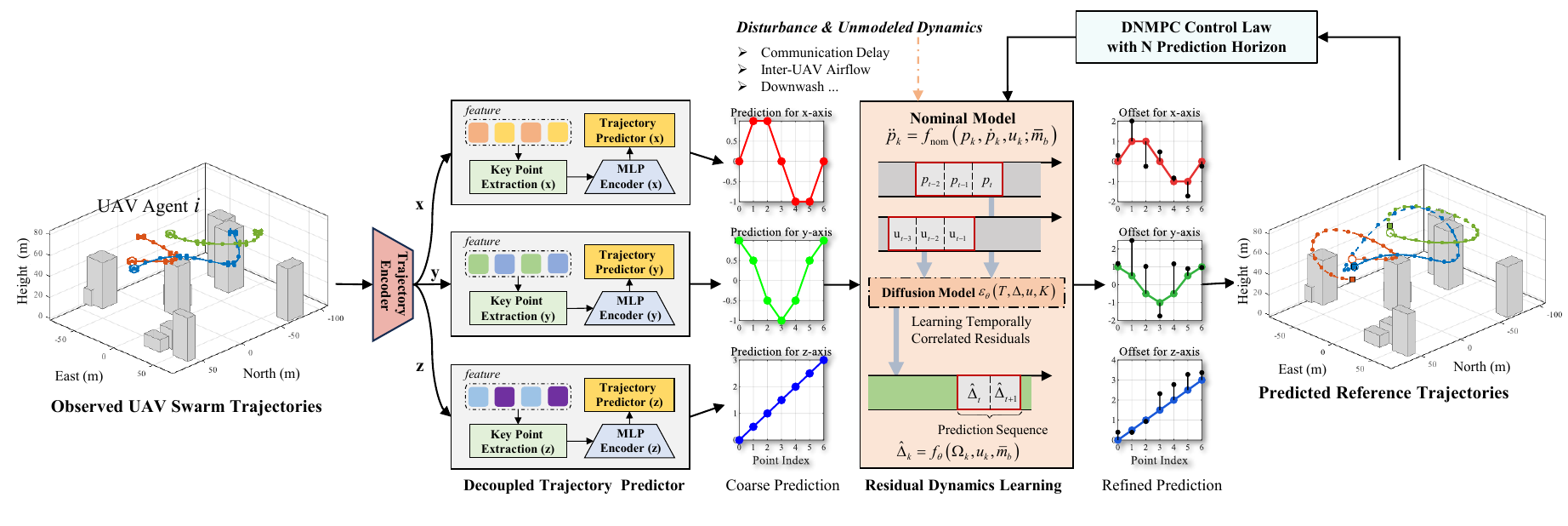}
    \caption{Framework of the proposed distributed trajectory prediction and control scheme for UAV swarms. Future trajectories are first predicted in a decoupled manner along each spatial axis from historical observations. A diffusion-based residual dynamics module then refines these predictions by compensating for unmodeled dynamics. The refined trajectories are incorporated into a DNMPC controller, where each UAV optimizes its control inputs over a finite horizon while satisfying inter-UAV formation constraints. This architecture enables accurate trajectory tracking and stable coordination under disturbances and modeling uncertainties.}
    \label{fig:overview}
\end{figure*}
\section{Related Work}

\subsection{Trajectory Prediction Methods}
Trajectory prediction aims to forecast the future motions of agents based on
their historical observations.
In multi-agent settings, a central challenge lies in modeling inter-agent
interactions and collective behaviors under uncertainty.
To address the inherent diversity and stochasticity of future trajectories,
a wide range of generative models has been proposed, including Generative
Adversarial Networks (GANs)~\cite{salzmann2020trajectronpp},
Conditional Variational Autoencoders (CVAEs)~\cite{yuan2021agentformer},
and diffusion-based models~\cite{bansal2019chauffeurnet,yang2025unified}.
Sequential models based on recurrent neural networks (RNNs)~\cite{Ivanovic2021HeterogeneousAgentTF}
and Transformers~\cite{VNAGT} further enable flexible handling of
variable-length inputs.
These approaches have demonstrated strong predictive performance on benchmark
datasets, particularly in pedestrian and vehicle trajectory prediction.

Despite their success, most existing trajectory prediction methods are designed
for offline forecasting or benchmarking purposes and primarily optimize
prediction accuracy without explicitly considering deployment in closed-loop
control systems.
As a result, their adaptability often degrades under real-world conditions
characterized by strong disturbances, sensing uncertainty, and complex agent
interactions.
In contrast, our work targets \emph{4D multi-UAV swarm trajectory prediction},
where temporal uncertainty, inter-agent coupling, and control-oriented
robustness must be jointly addressed.

\subsection{Diffusion Models for Motion Dynamics}

Diffusion models have recently emerged as a powerful paradigm for motion
generation across a variety of domains~\cite{Ho2020DenoisingDP}. Existing
diffusion-based approaches can be broadly categorized into two lines:
\emph{holistic trajectory diffusion} and \emph{end-to-end diffusion policy}
methods, which differ fundamentally in their modeling targets and system
integration.

\subsubsection{Holistic Trajectory Diffusion}
Holistic trajectory diffusion methods aim to model the entire spatiotemporal
trajectory as a single structured random variable, and directly learn its joint
distribution through diffusion processes. Representative works such as
SwarmDiff~\cite{Ding2025SwarmDiffSR} and MotionDiffuser~\cite{jiang2023motiondiffuser}
employ diffusion models to generate full future trajectories by capturing
high-order temporal correlations and cross-agent interactions in a unified
manner.

However, the holistic formulation often treats trajectory generation as an
open-loop process, without explicitly incorporating system dynamics or control
feedback during generation.

\subsubsection{End-to-End Diffusion Policy}

In contrast, end-to-end diffusion policy methods use diffusion models to directly
learn a mapping from observations to control actions, effectively treating
diffusion as a stochastic policy representation. In~\cite{Chi2023DiffusionPV}, the action sequence is generated directly
from sensory observations through a conditional diffusion process. Related works
in motion planning further integrate diffusion models with
reinforcement learning or cost-guided objectives~\cite{janner2022diffuser}, enabling the generation of
action or trajectory samples that optimize task-specific rewards or constraints~\cite{EDMP,zeng2025navidiffusorcostguideddiffusionmodel}.
Beyond robotics, diffusion-based policies have also demonstrated strong
performance in human motion synthesis, where high-dimensional motion sequences
are generated and edited under diverse conditioning signals, such as text
descriptions~\cite{Zhang2022MotionDiffuseTH} or partial state constraints~\cite{GuidedMotionDiffusion}.
While these end-to-end approaches offer seamless integration between perception
and control~\cite{HierarchicalDiffusionPolicy,Chi2023DiffusionPV}, they typically entangle dynamics modeling and decision-making into a
single learned policy, which can limit robustness in safety-critical control scenarios.

Motivated by these gaps, we propose a unified framework that couples
control-oriented 4D multi-UAV trajectory prediction with closed-loop tracking.
Specifically, we first generate coarse axis-wise forecasts to reduce prediction
complexity, then model temporally correlated inter-agent residual dynamics via
diffusion, and finally integrate the refined predictions into a DNMPC scheme, shown in Fig.~\ref{fig:overview}.
The proposed method is detailed in the following section.

\section{Method}

We consider a swarm consisting of $M$ UAVs,
indexed by $i\in\{1,\dots,M\}$.
Let $\mathbf{x}_{i,k}$ denote the true state of UAV $i$ at discrete time step $k$,
and $\bar{\mathbf{x}}_{i,k}$ the predicted state along the NMPC horizon.
We denote reference trajectories by
$\mathbf{x}^{\mathrm{ref}}_{i,k}$.
Position components are denoted by $p$, with $\bar{p}$ indicating predicted
positions and $\mathbf{p}^{obs}$ historical observations.
Each UAV maintains formation by enforcing inter-UAV distance constraints with
its neighbors.
Let $d^{\mathrm{ref}}_{ij}$ denote the desired relative distance between UAV $i$
and UAV $j$.
At time step $k$, the formation constraint is defined as:
\begin{equation}
\mathcal{C}_{ij}(k)
=
\big\|
\bar{p}_{i,k} - \bar{p}_{j,k}
\big\|_2
-
d^{\mathrm{ref}}_{ij},
\label{eq:formation_constraint}
\end{equation}
where $\bar{p}_{i,k}$ is extracted from $\bar{\mathbf{x}}_{i,k}$.

\subsection{Decoupled Trajectory Prediction}

Predicting future UAV trajectories in four-dimensional space requires
balancing model expressiveness and controllability.
Existing keypoint-conditioned methods such as LBEBM~\cite{Pang2021TrajectoryPW}
decode coupled 4D trajectories, which may entangle prediction errors across
spatial dimensions.
We therefore formulate trajectory forecasting as a
\emph{dimension-factorized prediction problem}.

We retain the encoder of LBEBM to extract interaction-aware representations from
historical observations $\mathbf{p}^{obs}_i$.
Let $\boldsymbol{\xi}_e$ denote the encoded latent feature.
Axis-specific key points are predicted as:
\begin{equation}
\mathbf{q}_{\kappa}^{\lambda} = \Phi_{\lambda}(\boldsymbol{\xi}_e),
\quad \lambda \in \{x,y,z\},
\end{equation}
where $\kappa$ indexes key points and $\Phi_\lambda(\cdot)$ is an
axis-dependent predictor.

The predicted key points are aggregated to form a compact structural descriptor:
\begin{equation}
\boldsymbol{\xi}_k =
\Psi\big([\mathbf{q}_{\kappa}^{x};\mathbf{q}_{\kappa}^{y};\mathbf{q}_{\kappa}^{z}]\big),
\end{equation}
where $\Psi(\cdot)$ encodes multi-axis geometric cues.
Trajectory decoding is then performed independently along each axis:
\begin{equation}
\hat{\mathbf{T}}^{\lambda}
=
\Gamma_{\lambda}\big([\boldsymbol{\xi}_e,\boldsymbol{\xi}_k]\big),
\quad \lambda \in \{x,y,z\}.
\end{equation}

\subsection{Residual Dynamics Refinement}

To compensate for modeling errors and aerodynamic disturbances, we introduce a
diffusion-based residual dynamics model.
The translational dynamics of a quadrotor are given by:
\begin{equation}
\dot{p}=v,\qquad
m_b\dot{v}=m_bg+Rf_u+f_a,
\end{equation}
where $m_b$ is the body mass, \(g\) is the gravity vector, and $f_a$ aggregates unmodeled disturbances.
Using a nominal mass $\bar{m}_b$, the dynamics are rewritten as:
\begin{equation}
\bar{m}_b\ddot{p}=u-\Delta(p,\dot{p},\ddot{p}),
\end{equation}
where $\Delta$ collects all state-dependent uncertainties. $\Delta(\cdot)$ is assumed to be an unknown but bounded and locally Lipschitz
residual function capturing unmodeled dynamics and aerodynamic disturbances.

We model residual dynamics over a finite estimation horizon $T$.
Define the observation set
$\Omega=\{p,\dot{p},\ddot{p}\}$ and the joint sequence:
\begin{equation}
\mathcal{S}=
[\Omega_{1:T},\,u_{1:T},\,\Delta_{1:T}]^\top.
\end{equation}
For each UAV, $p,\dot{p},\ddot{p} \in \mathbb{R}^3$, and the diffusion model is applied
agent-wise with shared parameters across the swarm.

Residual estimation is formulated as sampling from
$p(\Delta_{1:T}\mid\Omega_{1:T},u_{1:T})$.
The previous control input $u_{t-1}$ is used as a proxy under receding-horizon
operation. 

\subsection{DNMPC Control}

The refined residual estimate $\hat{\Delta}$ is incorporated into a DNMPC scheme.
We use $N$ to denote the NMPC prediction horizon, distinct from the residual
estimation horizon $T$. The residual horizon $T$ determines the temporal extent of uncertainty estimation, while the NMPC horizon 
$N$ governs control foresight, with $T \leq N$ ensuring consistent disturbance compensation within receding-horizon optimization.
Each UAV solves a local NMPC problem to track its reference trajectory while
respecting formation constraints.
The discrete-time prediction model is:
\begin{equation}
\bar{\mathbf{x}}_{i,k+j+1}
=
f_d(\bar{\mathbf{x}}_{i,k+j},\bar{\mathbf{u}}_{i,k+j},\hat{\Delta}_{i,k+j},\delta t).
\end{equation}
Only the first control input is applied at each step, yielding a receding-horizon closed-loop control law. 

\textbf{Proposition 1: Practical tracking stability.}
Assume recursive feasibility of the DNMPC subproblems. Then the optimal value function
$V_{i,k}:=J_i^\star(\mathbf{x}_{i,k})$ is Lyapunov-like for the closed-loop error dynamics, i.e.,
there exist $\alpha_i>0$ and $\beta_i>0$,
\begin{equation}
V_{i,k+1}-V_{i,k}
\le -\alpha_i \|\mathbf{e}_{i,k}\|_2^2
+ \beta_i\!\left(\bar{\delta}_\Delta^2+\bar{\delta}_n^2\right).
\label{eq:lyap_decrease}
\end{equation}
Consequently, the closed-loop tracking error is input-to-state stable w.r.t.
$(\tilde{\Delta}_{i,k},\,\bar{\mathbf{x}}_{j,k}-\mathbf{x}_{j,k})$ and is practically stable:
\begin{equation}
\limsup_{k\to\infty}\|\mathbf{e}_{i,k}\|_2
\le c_i \sqrt{\bar{\delta}_\Delta^2+\bar{\delta}_n^2},
\label{eq:practical_bound}
\end{equation}
for some constant $c_i>0$. In the ideal case $\tilde{\Delta}_{i,k}\equiv 0$ and $\bar{\delta}_n=0$,
\eqref{eq:lyap_decrease} implies asymptotic tracking of $\mathbf{x}^{\mathrm{ref}}_{i,k}$.

\begin{figure*}
    \centering
   \includegraphics[width=1\linewidth]{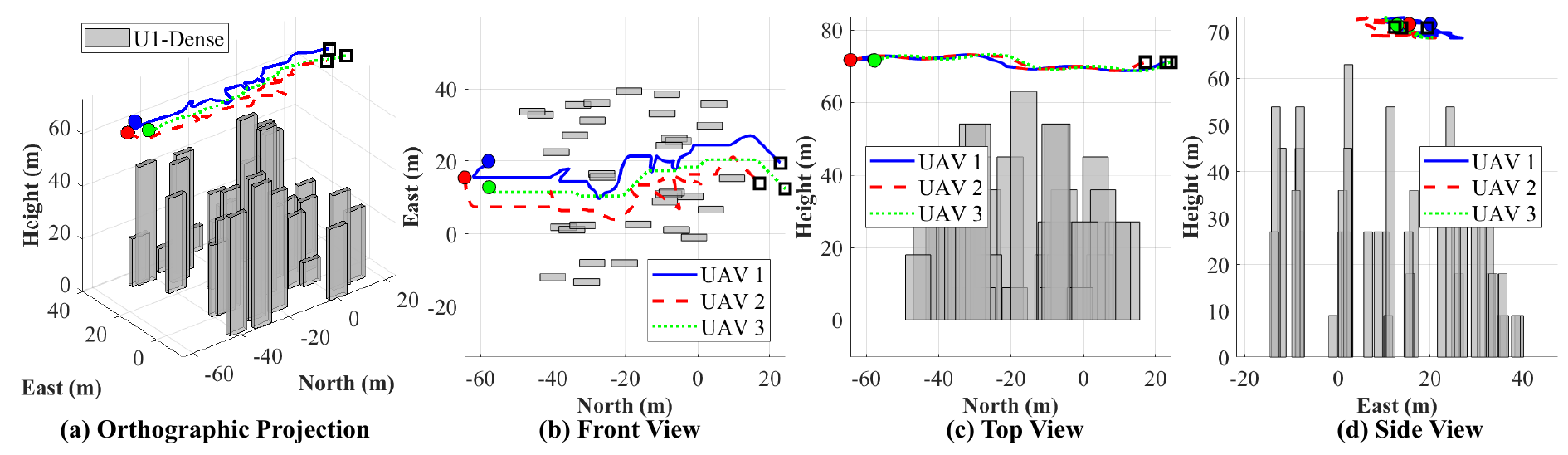}
    \caption{
    Visualization of real-world 4D UAV swarm trajectories in one urban scenario in the dataset.
    Each trajectory is parameterized by time and represented in three spatial dimensions (North, East, Height).
    The orthographic projection and the front, top, and side views correspond to different projections of the same time-indexed trajectories, illustrating the temporal evolution and spatial coordination of the UAV swarm. The curvature of these trajectories is caused by communication delay and external wind. 
    }
    \label{fig:TWO}
\end{figure*}

\section{4D Motion Trajectory Dataset}
Our dataset comprises six scenarios featuring 4D trajectories collected from UAVs operating in real typical urban environments. The 4D trajectories, shown in Fig.~\ref{fig:TWO}, are recorded using positioning devices installed on the UAVs. Each scenario includes three UAVs following predefined formations. 
The three-UAV setting is chosen to reflect tightly coupled formations, while the
distributed NMPC formulation does not impose an upper bound on swarm size.

The dataset contains a total of 7,900 frames, with scenario lengths ranging
from 800 to 2,000 frames.
As summarized in Table~\ref{tab:scenario_stats}, the inter-UAV distance varies between
$1.3\,\mathrm{m}$ and $5.8\,\mathrm{m}$ on average, reflecting realistic
formation maintenance under environmental constraints.
The recorded flight speed spans $2.4$--$5.6\,\mathrm{m/s}$, while acceleration
and turning intensity statistics reveal diverse maneuvering behaviors across
scenarios.
These characteristics jointly capture both smooth cruising and aggressive
directional changes that frequently arise in practical low-altitude UAV
operations.

\begin{table*}[t]

\caption{Scenario-wise trajectory statistics (mean $\pm$ std) and intention distributions for drone swarm prediction. U, L, F, and R denote \emph{Urban}, \emph{Lab}, \emph{Factory}, and \emph{Rural} motion scenarios, respectively. 
}
\centering
\small
\setlength{\tabcolsep}{6pt}
\renewcommand{\arraystretch}{1.25}
\rowcolors{3}{gray!12}{white}

\begin{tabularx}{\linewidth}{l *{6}{>{\centering\arraybackslash}X}}
\toprule
\rowcolor{gray!30}
\textbf{Scenario} & \textbf{U1-Dense} & \textbf{U2-Sparse} & \textbf{L1-Indoor} & \textbf{F1-Large} & \textbf{F2-Complex} & \textbf{R1-Open} \\
\midrule
Frame number
& 800 & 1600 & 1200 & 1500 & 2000 & 900 \\

Inter-UAV distance (m)
& $3.45\pm0.18$ & $5.80\pm0.22$ & $1.30\pm0.15$ & $1.70\pm0.25$ & $1.55\pm0.20$ & $3.90\pm0.30$ \\

Speed (m/s)
& $3.20\pm0.60$ & $4.10\pm0.70$ & $2.40\pm0.50$ & $4.80\pm0.80$ & $4.20\pm0.75$ & $5.60\pm0.90$ \\

Turning intensity (1/m)
& $0.18\pm0.10$ & $0.12\pm0.08$ & $0.22\pm0.12$ & $0.09\pm0.06$ & $0.15\pm0.09$ & $0.07\pm0.05$ \\

Speed intention
& \makecell{$\uparrow$:180,\\ $\downarrow$:260,\\ $\rightarrow$:360}
& \makecell{$\uparrow$:320,\\ $\downarrow$:420,\\ $\rightarrow$:860}
& \makecell{$\uparrow$:210,\\ $\downarrow$:290,\\ $\rightarrow$:700}
& \makecell{$\uparrow$:360,\\ $\downarrow$:420,\\ $\rightarrow$:720}
& \makecell{$\uparrow$:480,\\ $\downarrow$:520,\\ $\rightarrow$:1000}
& \makecell{$\uparrow$:200,\\ $\downarrow$:260,\\ $\rightarrow$:440} \\

Target sector 
& \makecell[tl]{\#1:90, \#2:80,\\ \#3:140, \#4:60,\\ \#5:120, \#6:110,\\ \#7:150, \#8:50}
& \makecell[tl]{\#1:160, \#2:140,\\ \#3:260, \#4:180,\\ \#5:300, \#6:280,\\ \#7:190, \#8:90}
& \makecell[tl]{\#1:110, \#2:160,\\ \#3:220, \#4:180,\\ \#5:90, \#6:150,\\ \#7:210, \#8:80}
& \makecell[tl]{\#1:200, \#2:220,\\ \#3:260, \#4:300,\\ \#5:180, \#6:220,\\ \#7:80, \#8:40}
& \makecell[tl]{\#1:260, \#2:300,\\ \#3:320, \#4:360,\\ \#5:240, \#6:260,\\ \#7:180, \#8:80}
& \makecell[tl]{\#1:120, \#2:140,\\ \#3:160, \#4:180,\\ \#5:110, \#6:100,\\ \#7:60, \#8:30} \\
\bottomrule
\end{tabularx}

\label{tab:scenario_stats}
\end{table*}

\section{Experiments}

\subsection{Implementation Details}

We consider a short-horizon trajectory prediction setting, where the past
8 frames are used as observations to forecast the subsequent 12 frames.
Rather than predicting all future states densely, the model outputs a set of
representative future key points corresponding to the 3rd, 6th, 9th, and 12th
prediction steps, which are subsequently interpolated and incorporated into
the NMPC reference trajectory.
This design reduces prediction redundancy while preserving sufficient
temporal resolution for control.

Model training is performed for 100 epochs with a batch size of 70 using the
Adam optimizer.
The initial learning rate is set to $5\times10^{-4}$ and decayed by a factor
of 0.5 after the fifth epoch.
Prediction accuracy is evaluated using Average Displacement Error (ADE) and
Final Displacement Error (FDE), while closed-loop control performance is
assessed using trajectory tracking error and RMSE.

The diffusion noise predictor $\epsilon_{\theta}$ is implemented using a
temporal U-Net architecture composed of three residual stages~\cite{janner2022diffuser}, each including
two temporal convolution layers followed by group normalization~\cite{Wu2018GroupN} and Mish
activation~\cite{Misra2020MishAS}.
To encode diffusion timesteps and conditional inputs, we employ two separate
two-layer MLPs with 64 hidden units, producing 16-dimensional embeddings.
The noise predictor is optimized with Adam using a learning rate of
$2\times10^{-4}$ over 50{,}000 training steps with a batch size of 256 and
$K=20$ diffusion steps.

For control performance evaluation, we benchmark the proposed framework against a physics-driven DNMPC baseline, which has demonstrated strong stability and coordination capabilities in multi-agent control scenarios~\cite{stomberg2024decentralized}. To assess the benefit of incorporating diffusion-based residual modeling, we further compare it with representative data-driven dynamics learning methods, including Gaussian Processes (GP) models~\cite{torrente2021data} and deep neural network (DNN) predictors~\cite{shi2019neural}, both of which have been widely adopted for quadrotor dynamics approximation. An autoregressive sequence modeling baseline based on GPT4~\cite{liu2025indooruavbenchmarkingvisionlanguageuav} is also included. The GPT-based autoregressive model serves as a strong sequence modeling baseline to assess whether generic temporal predictors can approximate residual dynamics. For a fair tracking comparison, all learning-based models are trained on the same offline dataset to predict residual dynamics $\Delta$, and are integrated into the identical control law in~(8). Each method is evaluated over 10 independent trials.

\subsection{Comparisons to State-of-the-art Methods}
We compare our method with representative baselines spanning classical filtering, deterministic sequence models, and generative trajectory predictors, as detailed in Table~\ref{tab:efficiency_accuracy}. Kalman Filter (KF) and Unscented Kalman Filter (UKF) serve as classical model-based baselines for state estimation~\cite{UKF}. Temporal Convolutional Networks (TCN) and MSRL represent deterministic learning-based predictors~\cite{BaiTCN2018,wu2023multi}. LBEBM and NPSN capture energy-based and probabilistic sequence modeling paradigms~\cite{Pang2021TrajectoryPW,bae2022npsn}. GAN- and CVAE-based methods are included as canonical generative baselines for multimodal trajectory prediction~\cite{SocialBiGAT,zhou2022dynamic}, while MTRTraj represents recent large-capacity transformer-based trajectory models~\cite{shi2022motion}.

We modify all compared methods to accept 4D observed trajectories as inputs and predict 4D future trajectories. Additionally, we adjust their data augmentation strategies, such as rotation, flipping, and reversing, to their 4D counterparts for consistency. The experimental results indicate that our method outperforms all evaluated approaches in 4D trajectory prediction. Specifically, our method achieves a 16.7\% improvement in the ADE metric, reducing it from 0.18 to 0.15, and a 18.2\% enhancement in the FDE metric, lowering it from 0.33 to 0.27, compared to the second-best method, MTRTraj. While Unscented Kalman Filter (UKF) shows an ADE metric close to ours, our FDE metric surpasses it by 48\%, decreasing from 0.52 to 0.27.

Table~\ref{tab:efficiency_accuracy} also compares prediction accuracy and computational efficiency across representative baselines. Classical filters (KF/UKF) achieve low latency but suffer from large prediction errors, while lightweight neural models such as TCN and MSRL offer moderate accuracy gains at increased cost. Our diffusion-based prediction improves accuracy but incurs substantial runtime overhead due to iterative sampling. Our method adopting DDIM acceleration~\cite{Song2020DenoisingDI} attains the best accuracy-efficiency trade-off, achieving the lowest ADE/FDE with sub-30 ms inference latency at about 34 FPS. These results indicate that reusing multistep diffusion predictions effectively reduces sampling overhead while preserving real-time feasibility for control deployment.

\begin{table}[t]
\caption{Trajectory prediction efficiency and accuracy comparison on the dataset.
All methods are evaluated on an NVIDIA RTX 4070 GPU with an input size of $70 \times 8 \times 3$, where 70 denotes the number of agents predicted simultaneously, exceeding the scale of most real-world deployment scenarios. KF-based methods do not involve learnable parameters, so parameter counts are omitted.}
\centering
\small
\setlength{\tabcolsep}{3.5pt}
\renewcommand{\arraystretch}{1.18}
\begin{tabularx}{\linewidth}{X c c c c c}
\toprule
\multirow{2}{*}{\textbf{Method}} 
& \multicolumn{3}{c}{\textbf{Efficiency}} 
& \multicolumn{2}{c}{\textbf{Accuracy \(\Downarrow\)}} \\
\cmidrule(lr){2-4} \cmidrule(lr){5-6}
& Params (M) & FLOPs (G) & Speed (ms) & ADE  & FDE  \\
\midrule
KF          
& -- & \textbf{0.08}  & 18 & 0.31 & 0.58 \\
UKF         
& -- & 0.11  & 22 & 0.18 & 0.52 \\
TCN         
& \textbf{0.21} & 0.18  & 26 & 0.24 & 0.43 \\
MSRL        & 0.62 & 0.14  & 20 & 0.23 & 0.41 \\
LBEBM       & 1.31 & 0.12  & \textbf{15} & 0.20 & 0.36 \\
NPSN        & 0.28 & 0.19  & 29 & 0.21 & 0.38 \\
GAN         & 0.44 & 0.24  & 31 & 0.22 & 0.39 \\
CVAE        & 0.58 & 0.27  & 34 & 0.20 & 0.36 \\
MTRTraj    & 4.52 & 6.80  & 34 & 0.28 & 0.33 \\

\rowcolor{gray!15}
\textbf{Ours} 
            & 3.48 & 0.42  & 48 & 0.17 & 0.31 \\
            \rowcolor{gray!15}
\textbf{w DDIM} 
            & 3.48 & 0.26 & 29 
            & \textbf{0.15} & \textbf{0.27} \\          
\bottomrule
\end{tabularx}
\label{tab:efficiency_accuracy}
\end{table}

Thus, the diffusion-based predictor provides NMPC with more reliable and temporally consistent state rollouts, reducing abrupt prediction errors that may otherwise lead to infeasible or aggressive control actions. This effect becomes particularly pronounced under tight formation constraints and high-speed maneuvers, where small modeling errors can be significantly amplified by inter-UAV interactions. As a result, DM-NMPC achieves improved tracking accuracy and robustness by implicitly regularizing the NMPC optimization landscape, leading to smoother control inputs and enhanced stability in multi-UAV swarm.

\subsection{Robustness and Adaptation Analysis}

Table~\ref{tab:model_tracking_compact} compares learning-assisted NMPC controllers across four motion scenarios. Overall, DM-NMPC achieves the lowest prediction and tracking errors, indicating improved compensation for unmodeled dynamics and inter-UAV interactions. The prediction error, measured in Newtons, quantifies the accuracy of estimating
interaction-induced and environment-dependent disturbance forces between neighboring UAVs.
Accurate force prediction is critical for DNMPC, as these estimates are directly
used in the feedback compensation term of the control law.
Compared with GP-NMPC and DNN-NMPC, our DM-NMPC reduces trajectory tracking RMSE by 15-22\% in Urban and Factory scenarios and by 10-15\% in Lab and Rural environments. 
\begin{table}[t]
\caption{Force prediction and tracking errors of different DNMPC-based controllers. 
}
\centering
\footnotesize
\setlength{\tabcolsep}{3pt}
\renewcommand{\arraystretch}{1.15}

\begin{tabular}{l cccc cccc}
\toprule
& \multicolumn{4}{c}{\textbf{Pred. Err. (N) \(\Downarrow\)}} 
& \multicolumn{4}{c}{\textbf{Track. Err. (m) \(\Downarrow\)}} \\
\cmidrule(lr){2-5} \cmidrule(lr){6-9}
\textbf{NMPC} \\ Method 
& U & L & F & R 
& U & L & F & R \\
\midrule
GP
& 0.139 & 0.147 & 0.105 & 0.116
& 0.091 & 0.096 & 0.076 & 0.071 \\

DNN
& 0.128 & 0.140 & 0.096 & 0.108
& 0.081 & 0.089 & 0.069 & 0.064 \\

GPT
& 0.121 & 0.129 & 0.091 & 0.099
& 0.075 & 0.083 & 0.063 & 0.059 \\

\textbf{DM}
& \textbf{0.104} & \textbf{0.112} & \textbf{0.079} & \textbf{0.089}
& \textbf{0.066} & \textbf{0.071} & \textbf{0.055} & \textbf{0.052} \\
\bottomrule
\end{tabular}

\label{tab:model_tracking_compact}
\end{table}

\begin{figure}
    \centering
    \includegraphics[width=1\linewidth]{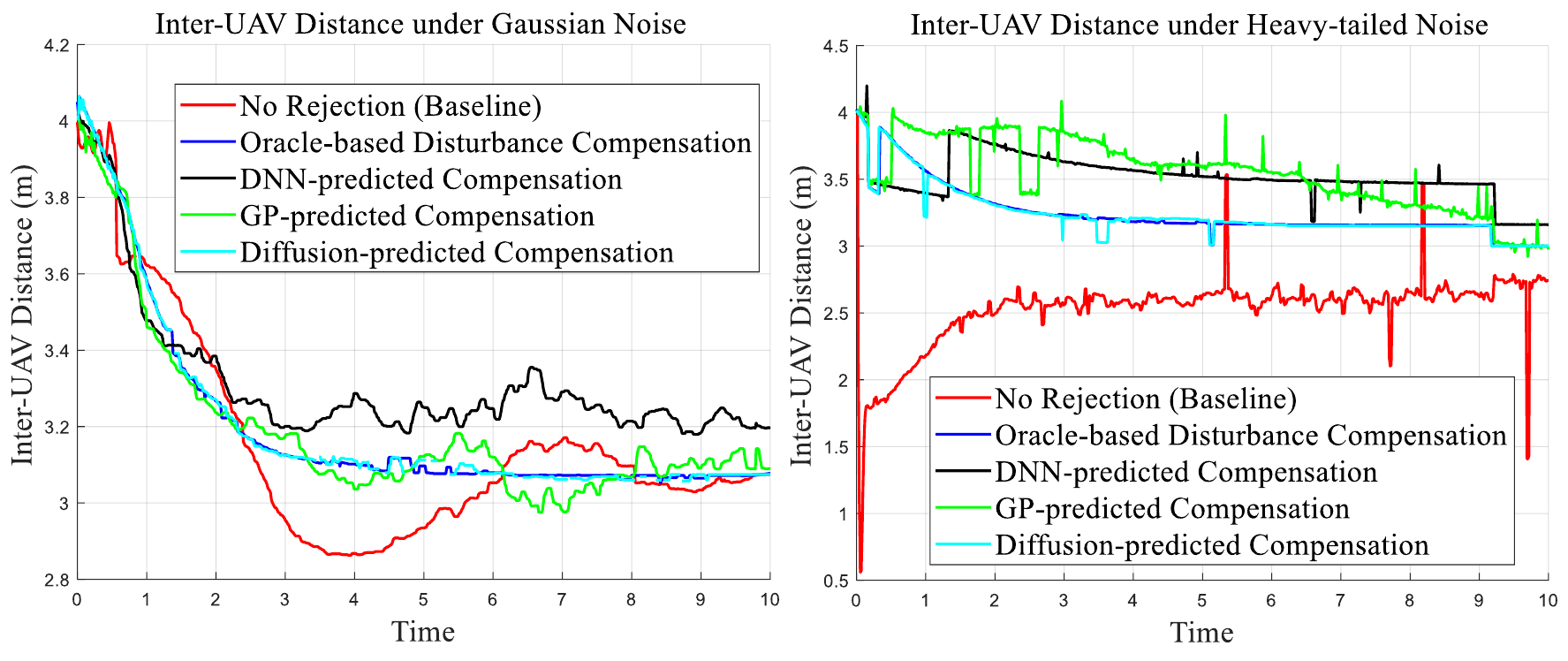}
    \caption{Convergence comparison of the inter-UAV distance dynamics under different disturbance compensation strategies, evaluated with 10 s of training data sampled at 20 Hz.}
    \label{fig:NoiseCompare}
\end{figure}

Fig.~\ref{fig:NoiseCompare} compares inter-UAV distance convergence under Gaussian (left) and heavy-tailed (right) disturbances. Under Gaussian noise, most learning-based compensation strategies achieve stable convergence, with GP-based methods benefiting from smoothness assumptions. However, under heavy-tailed Cauchy disturbances, GP- and DNN-based compensators exhibit degraded performance, characterized by persistent oscillations or steady-state offsets. In contrast, diffusion-based compensation consistently maintains bounded inter-UAV distance with reduced oscillations across both regimes, demonstrating superior robustness to non-Gaussian and extreme disturbances in multi-UAV formation control.

Under different reference spacing and speed settings, Table~\ref{tab:robustness} reports the formation constraint RMSE of various NMPC variants. When the speed is increased from 8 m/s to 10 m/s, the RMSE of all methods increases. The baseline methods exhibit a more pronounced increase in error under the combination of high speed and large spacing, reflecting their limited ability to compensate for unmodeled disturbances or residual dynamics. In contrast, DM-NMPC achieves the lowest RMSE across all conditions and maintains a relatively stable error level even in the more challenging setting (10 m/s, 0.8 m), indicating its capacity to more effectively absorb prediction errors and disturbances during the closed-loop control process, thereby enhancing the robustness and consistency of the formation constraints.

\begin{table}[t]
\caption{Formation-keeping RMSE of different NMPC variants under varying reference distances and flight velocities.}
\centering
\small
\setlength{\tabcolsep}{4pt}
\renewcommand{\arraystretch}{1.2}

\begin{tabular}{l cccc c}
\toprule
\multirow{2}{*}{\textbf{Method}} 
& \multicolumn{4}{c}{\textbf{Formation Constraint RMSE (m) \(\Downarrow\)}} 
 \\
\cmidrule(lr){2-5}
& \multicolumn{2}{c}{$d_{ij}^{ref}{=}1.2$m} 
& \multicolumn{2}{c}{$d_{ij}^{ref}{=}0.8$m} 
&  \\
& $v{=}8$ m/s
& $v{=}10$ m/s
& $v{=}8$ m/s
& $v{=}10$ m/s
&  \\
\midrule
DNMPC
& 0.118 & 0.146 & 0.156 & 0.181 \\

GP-NMPC
& 0.123 & 0.152 & 0.162 & 0.189 \\

DNN-NMPC
& 0.109 & 0.135 & 0.144 & 0.168 \\

GPT-NMPC
& 0.101 & 0.124 & 0.133 & 0.156 \\

\textbf{DM-NMPC}
& \textbf{0.089} & \textbf{0.107} & \textbf{0.118} & \textbf{0.137} \\

\bottomrule
\end{tabular}

\vspace{1mm}
\label{tab:robustness}
\end{table}

\subsection{Prediction Efficiency and Runtime Performance}

\begin{figure}[t]
    \centering
    \includegraphics[width=1\linewidth]{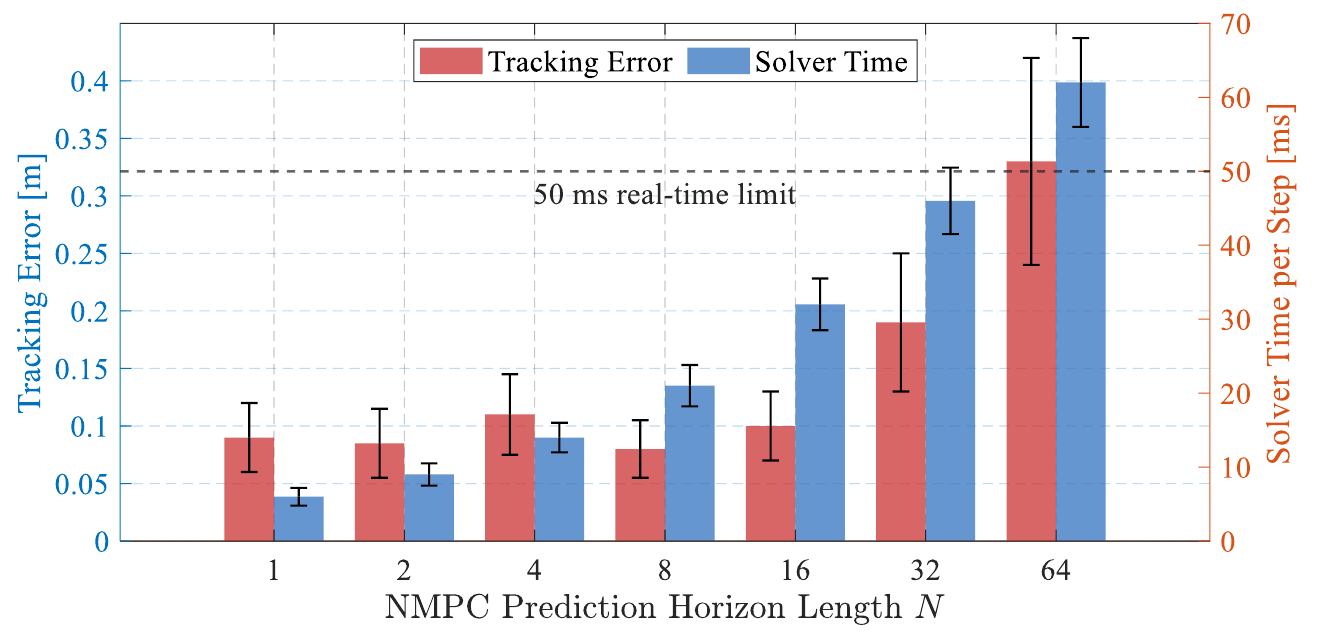}
    \caption{Steady-state tracking error and solver time during a hovering task
under different NMPC prediction horizons N over 60 seconds.}
    \label{fig:track}
\end{figure}

As illustrated in Fig.~\ref{fig:track}, increasing the prediction horizon initially improves tracking accuracy because the controller can anticipate future motion more effectively. However, when the horizon becomes too long, performance deteriorates as modelling errors accumulate and the system is more likely to violate real-time assumptions. In parallel, the solver time per step rises almost linearly with the horizon length and exceeds the 50~ms real-time limit when \(N \ge 32\). Overall, a moderate horizon of \(N\), from 8 to 16, offers the most balanced compromise between tracking quality and computational feasibility for DNMPC in UAV swarms. This decline at large \(N\) is mainly due to estimating \(\Delta\) from increasingly outdated state observations and control inputs.

\begin{figure}[h]
    \centering
    \includegraphics[width=0.98\linewidth]{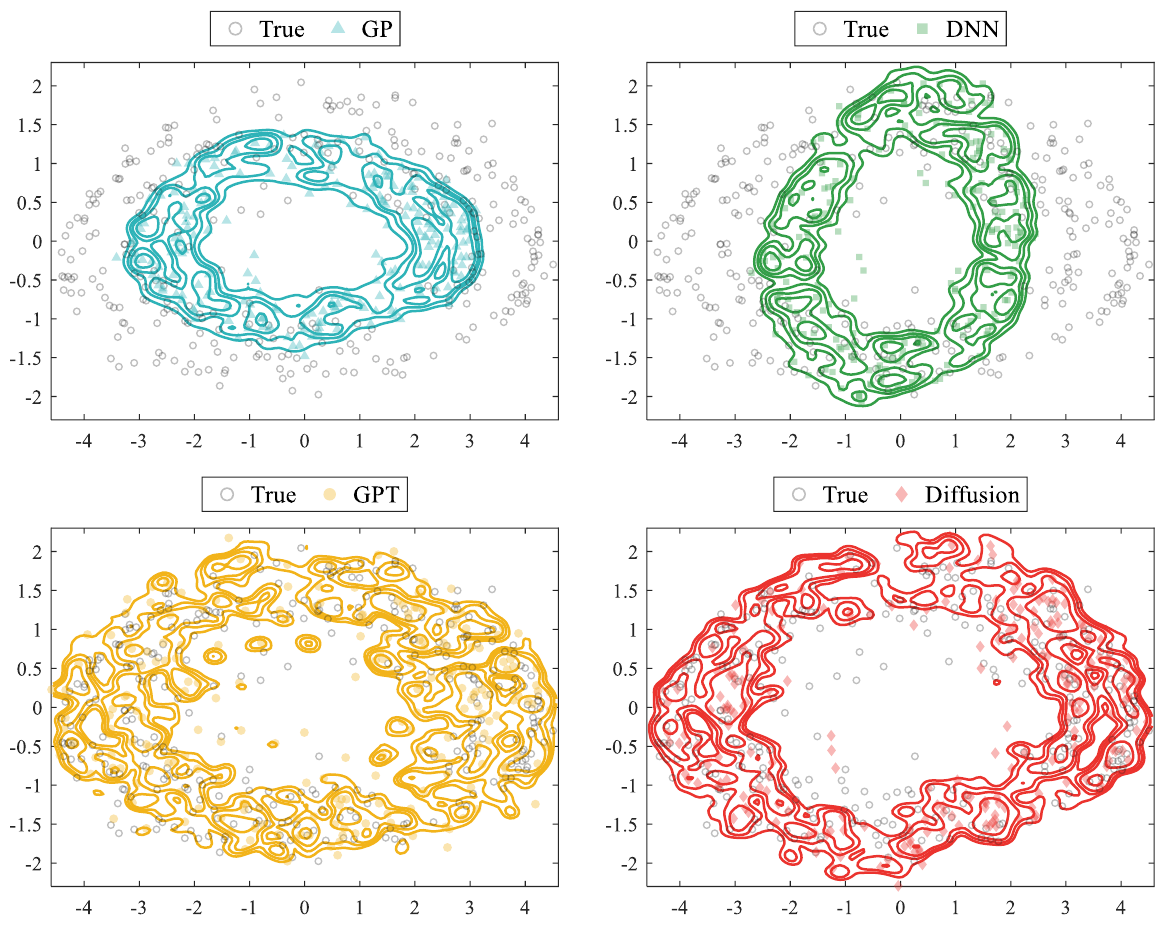}
    \caption{
    t-SNE visualization of residual dynamics $\Delta$ extracted from real-world flight data (red circle in Fig.~1 (b)) and estimated by different models. Compared with GP-, DNN-, and GPT-based predictors, the proposed diffusion model exhibits a distribution that more closely aligns with the true residual manifold.
}
    \label{fig:tSNE}
\end{figure}

As shown in Fig.~\ref{fig:tSNE}, GP and DNN tend to concentrate in the inner region or exhibit distorted coverage, while GPT produces scattered interior mass. In contrast, the diffusion-based model aligns well with the high-density manifold, particularly along the outer ring, indicating improved distributional fidelity.

\subsection{Failure Case Analysis}
\begin{figure}[t]
    \centering
    \includegraphics[width=1\linewidth]{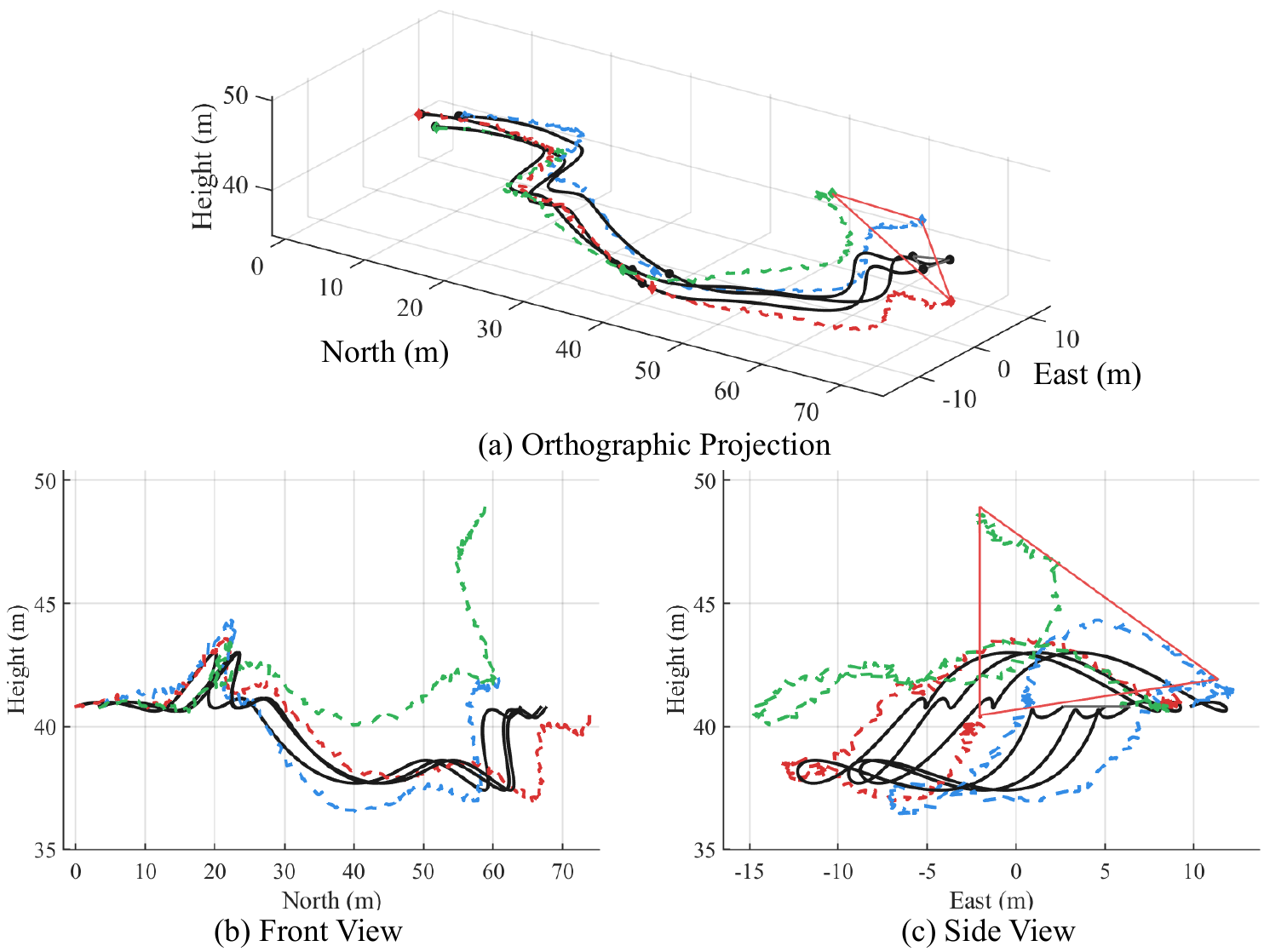}
    \caption{A representative failure case under strong inter-UAV interaction and highly curved motion, where accumulated residual prediction errors lead to formation degradation. Black solid lines denote ground-truth trajectories, while dashed lines indicate failed predictions.}
    \label{fig:Failcase}
\end{figure}

We further analyze a representative failure case in Fig.~\ref{fig:Failcase}.
Although our method accurately captures short-term motion trends, prediction errors accumulate in highly curved segments where inter-UAV interactions intensify.
This suggests that the current residual dynamics model, trained under fixed formation assumptions, may be insufficient to fully capture abrupt interaction changes or rare coordination patterns.
We leave adaptive interaction modeling and uncertainty-aware horizon adjustment as future work.

\subsection{Generalization Discussion}
To evaluate cross-dataset generalization, we additionally conduct trajectory
prediction experiments on the public CityNav dataset~\cite{lee2025citynavlargescaledatasetrealworld}. Since CityNav does not
provide synchronized multi-UAV trajectories or control annotations, this
evaluation focuses on open-loop trajectory prediction rather than closed-loop
control. As shown in Table~\ref{tab:citynav_generalization}, despite the domain gap between datasets, our method achieves consistently lower
prediction errors, suggesting enhanced robustness to scenario variations.

\begin{table}[t]
\centering
\small
\setlength{\tabcolsep}{6pt}
\renewcommand{\arraystretch}{1.2}
\caption{Cross-dataset trajectory prediction performance on the CityNav dataset.
Models are trained on our dataset and evaluated on CityNav without retraining.
}
\label{tab:citynav_generalization}
\begin{tabular}{lcc}
\toprule
\textbf{Method} & \textbf{ADE (m)} $\Downarrow$ & \textbf{FDE (m)} $\Downarrow$ \\
\midrule
Kalman Filter      & 1.32 & 2.08 \\
LSTM-based      & 1.21 & 1.98 \\
Transformer-based  & 1.08 & 1.76 \\
Diffusion    & 0.97 & 1.62 \\
\midrule
\textbf{Ours}            & \textbf{0.86} & \textbf{1.44} \\
\bottomrule
\end{tabular}
\end{table}

\section{Conclusion and Future Work}
In this paper, we present a unified framework for 4D UAV swarm trajectory prediction and distributed control. By decoupling the trajectory prediction task and employing a diffusion model solely for residual dynamics refinement, our approach balances uncertainty modeling with the real-time demands of onboard control. 
Experimental results confirm that our method outperforms existing transformer and generative baselines. The distributed DNMPC formulation does not impose a fixed upper bound on the swarm size, and the residual model shares parameters across UAVs. However, the current dataset and empirical validation are limited to three-UAV formations, scaling to larger swarms remains future work.

\section{Acknowledgement} 
This work was partially supported by the 
National Natural Science Foundation of China Grant 62373315 and U24A20252, and National Key Research and Development Program of China Grant 2024YFB4707603.

\bibliographystyle{IEEEtran}

\bibliography{references}

\end{document}